\pdfoutput=1

\documentclass[11pt]{article}

\usepackage[]{EMNLP2023}

\usepackage{times}
\usepackage{latexsym}
\usepackage{graphicx}
\usepackage{amsmath}
\usepackage{amsfonts}
\usepackage{multirow}
\usepackage{changepage}
\usepackage{pifont}
\usepackage[T1]{fontenc}

\usepackage[utf8]{inputenc}

\usepackage{microtype}

\usepackage{inconsolata}

\newcommand{\cmark}{\ding{51}}%
\newcommand{\xmark}{\ding{55}}%

\title{\textsc{HaLo:} Estimation and Reduction of Hallucinations in Open-Source Weak Large Language Models}

\author{
  Mohamed Elaraby \thanks{ 
Work conducted during an internship at TikTok US.} \textsuperscript{1} \quad Mengyin Lu \textsuperscript{2} \quad Jacob Dunn \textsuperscript{2} 
  \textbf{Xueying Zhang \textsuperscript{2}} \quad \\ \textbf{Yu Wang \textsuperscript{2}} \quad \textbf{Shizhu Liu \textsuperscript{2}} \quad  \textbf{Pingchuan Tian \textsuperscript{2}} \quad \textbf{Yuping Wang \textsuperscript{2}} \quad \textbf{Yuxuan Wang \textsuperscript{2}} \\
  \textsuperscript{1} University of Pittsburgh, Pittsburgh, PA, USA \\
  \textsuperscript{2} TikTok Inc., San Jose, CA, USA \\
  {\tt mse30@pitt.edu} \\ {\tt \{mengyin.lu,denglelai,xueying.zhang,yuwang.0827,shizhu.liu\}@bytedance.com} 
}

\begin{document}
\maketitle
\begin{abstract}

Large Language Models (LLMs) have revolutionized Natural Language Processing (NLP). Although convenient for research and practical applications, open-source LLMs with fewer parameters often suffer from severe hallucinations compared to their larger counterparts. This paper focuses on measuring and reducing hallucinations in \textit{BLOOM 7B}, a representative of such weaker open-source LLMs that are publicly available for research and commercial applications. We introduce \textsc{HaloCheck}, a lightweight BlackBox knowledge-free framework designed to quantify the severity of hallucinations in LLMs. Additionally, we explore techniques like \textit{knowledge injection} and \textit{teacher-student} approaches to alleviate hallucinations in low-parameter LLMs.
Our experimentation, carried out on a domain-specific question answering dataset centered around the NBA domain, underscores the efficacy of our proposed methods in effectively estimating and reducing hallucinations in the generated output.
 \url{https://github.com/EngSalem/HaLo}


\end{abstract}

\section{Introduction}

Despite their remarkable capabilities in various natural language processing tasks, LLMs often generate inaccurate or misleading information, leading to hallucination problems \cite{dong2022survey, min2023factscore, manakul2023selfcheckGPT}. These hallucinations can manifest as the generation of fictional or fabricated details, making it difficult to rely on LLM outputs for critical applications. While previous research on hallucination mitigation and detection has primarily focused on LLMs with parameters exceeding 100B parameters (such as GPT) \cite{brown2020language, ouyang2022training, bubeck2023sparks}, limited work has addressed smaller LLMs, which often lag behind larger models in performance \cite{wang2023aligning, wang2023far}. 
\textit{In this paper, we specifically address the estimation and mitigation of hallucinations in an open-source LLM with fewer parameters, namely BLOOM 7B \cite{scao2022bloom}, which is publicly available for both academic and commercial purposes}s \footnote{Stronger open-source LLMs have become available for both commercial  and academic use during our research such as LLama2 \cite{touvron2023llama}. We consider them as potential extensions for future work.}. 

\begin{figure*}[h]
\small
\begin{center}

 \includegraphics[width=16.4cm,height=4cm]{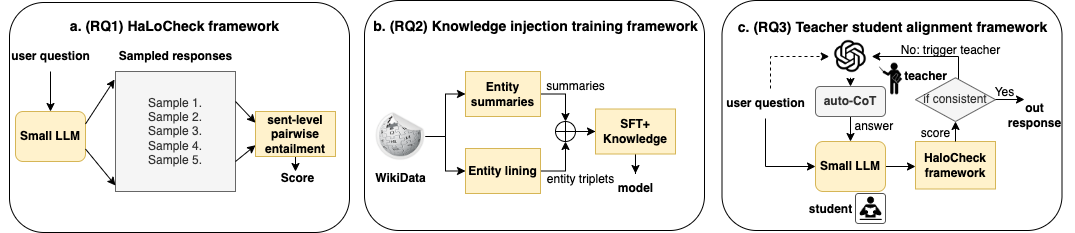}
 \caption{\label{halo_frame} The \textsc{Halo} framework is illustrated in three parts. (a) \textsc{HaloCheck} framework components for hallucinations estimation. (b) Knowledge injection training, which aims to reduce hallucinations by injecting abstract knowledge representations. (c) Improving student responses by stronger teacher responses, wherein teacher involvement is selectively triggered (indicated by the $--$ line) to manage the frequency of teacher involvement.}
 \end{center}
\end{figure*}

Our work aims to address three research questions. \textbf{RQ1: How can we quantify the severity of hallucinations in LLMs?}
In the context of addressing hallucination reduction, our initial focus is on developing an automated and efficient method to estimate the severity of hallucinations present in model-generated outputs. To tackle this, we introduce \textsc{HaloCheck}, a lightweight sampling-based BlackBox knowledge-free framework. This framework employs sentence-level entailment techniques to quantitatively assess the extent of hallucination within the generated output (as illustrated in Figure \ref{halo_frame} a.). Our approach aligns with other similar sampling-based BlackBox metrics \cite{manakul2023selfcheckGPT}, but also addresses their limitations in accurately estimating the severity of hallucinations.
\textbf{RQ2: How can we enhance the knowledge of LLMs without resorting to instruction tuning?}  We aim to enhance the knowledge of smaller LLMs through  \textit{Knowledge Injection (KI)}. This method involves fine-tuning the small LLM with domain-specific knowledge, without relying on manual instructions or instructions  obtained from stronger LLMs, which can be resource-intensive to generate at scale.  Additionally, instructions obtained from stronger LLMs might not be usable outside of research contexts \footnote{For example, Alpacha instructions are restricted to use beyond research purposes.}. We explore two avenues for knowledge injection: \textit{entity summaries} and \textit{entity triplets} (as shown in Figure \ref{halo_frame} b.), sourced from the WikiData knowledge base \cite{vrandevcic2014wikidata}. Further details about this approach can be found in Section \ref{sec:k_i}. \textbf{RQ3: Can a more robust teacher LLM helps guiding weaker LLM?} We investigate leveraging a more powerful LLM, specifically focusing on GPT-4, to provide guidance to weaker LLMs by generating detailed answers to questions. We also examine the practical use of \textsc{HaloCheck} as a hallucination estimator. Through assessing the severity of hallucination, we aim to optimize the engagement of the teacher LLM, thereby reducing the need for frequent queries to the teacher model (Figure \ref{halo_frame} c.). This strategy offers the advantage of alleviating the computational load and costs associated with heavy reliance on extensive-parameter LLMs.

\section{Related Work}

\paragraph{Hallucinations in LLMs.}
LLMs undergo pretraining through autoregressive methods, aiming to maximize the probability $P(w|c)$ of predicting the subsequent word(s) $w$ given the previous context $c$. This process equips the model with language understanding, knowledge, and reasoning capabilities \cite{jawahar-etal-2019-bert,raffel2020exploring,kojima2022large}. Based on this objective,  \citet{manakul2023selfcheckGPT} suggested that hallucinations are likely to occur when the model generates word(s) $w$ that deviate from the optimal prediction due to flat probabilities in certain contexts. This hypothesis suggests that hallucinations may emerge when the model lacks clear and confident predictions, leading to the generation of less accurate or irrelevant information. \textit{Inspired by this hypothesis, we propose \textsc{HaloCheck}, a lightweight BlackBox knowledge-free framework for evaluating hallucinations severity in LLMs. By sampling multiple responses to a given prompt and assessing their fine-grained agreement, \textsc{HaloCheck} helps identify and measure the model's tendency to generate consistent answers.}

\paragraph{Hallucinations detection in LLMs} 
Detecting hallucinations has been explored across various NLP tasks, including
summarization \cite{tam-etal-2023-evaluating}, dialogue \cite{shuster2021retrieval}, and question answering \cite{nori2023capabilities}. Recent work in detecting hallucinations of LLMs can be categorized into two approaches: \textit{automatic evaluation} and \textit{human evaluation} \cite{min2023factscore,zhang2023language}. \textit{In this work, we focus on automatic evaluation to reduce the reliance on human annotators.} Closely related to our work, \cite{manakul2023selfcheckGPT}  introduced \textit{selfcheckGPT}, a BlackBox knowledge-free metric that leverages existing evaluation measures, including BERTScore \cite{zhang2019bertscore}, and QA-based evaluation metrics \cite{manakul2023mqag}, and n-gram language models \footnote{selfcheckGPT-NLI was introduced shortly before the release of our work , yet it  wasn't included in the original paper.}, to measure the agreement between generated samples in natural language generation tasks. Similarly, \citet{mundler2023self}  introduced an LLM-based BlackBox evaluation metric to measure contradictions within generated samples in text generation tasks. This metric utilizes the LLMs' capacity to assess the contradictions present in the generated text. \textit{Our proposed \textsc{HaloCheck} framework is  another addition to the BlackBox evaluation category. We build on top of a lightweight entailment-based approach, \textsc{SummaC} \cite{laban-etal-2022-summac}, which was mainly designed to detect hallucinations in generated summaries, to detect sentence-level contradictions in generated responses, avoiding costly LLM calls. It addresses the limitations of selfcheckGPT by offering a more nuanced and efficient evaluation of severity of hallucinations in LLMs.} 

\paragraph{Hallucinations reduction in LLMs}

Previous research has focused on mitigating LLM hallucinations through augmented retrieval techniques \cite{mialon2023augmented}. For instance, REALM \citet{guu2020retrieval} retrieves relevant documents to prompts from external knowledge bases to enhance question answering performance in LLMs. \citet{lazaridou2022internet} used Google search results to improve question answering with extensive knowledge exploration. \citet{zhao2023verify} introduced "verify and edit" framework that first checks the generated response and edits unfactual parts using retrieved knowledge. Similarly, Chain-of-Knowledge \citet{li2023chain} employs knowledge triplets from external sources to enhance factuality in generated answers.
\textit{Our research focuses on integrating external knowledge sources directly into the model through Supervised Finetuning (SFT). This approach aims to enhance the base model's understanding and knowledge from the outset, reducing the reliance on constant external knowledge queries during inference. By doing so, we strive to bolster the model's knowledge and generation abilities without the need for instructions manually created or obtained from a stronger model.}



\paragraph{Teacher student approaches}
Knowledge distillation from larger models has become a prevalent technique in deep learning research \cite{hinton2015distilling}. Recent efforts, like Alpacha \cite{bommasani2021opportunities}, have explored distilling knowledge from teacher Language Models (LLMs) to student LLMs through finetuning, using instructions derived from the teacher model. Similarly, other studies have attempted to enhance student LLMs' explanations by finetuning them with explanation instructions obtained from more robust teacher models \cite{magister2022teaching,fu2023specializing,ho2022large}. 
\citet{saha2023can} proposed a method of teacher interventions to enhance reasoning capabilities in weaker Language Models (LLMs). In this approach, a teacher LLM plays a crucial role in improving the weaker LLM by providing explanatory answers through the use of Theory of Mind (ToM) prompts \cite{moghaddam2023boosting}. By incorporating ToM prompts, the teacher LLM aids the weaker LLM in acquiring a deeper understanding and enhancing its reasoning abilities. \textit{We use teacher LLM guidance to mitigate hallucinated output in weaker students by offering detailed answers during inference when hallucinations arise. We utilize the \textsc{HaLoCheck} metric (RQ1) to prompt the teacher's intervention selectively. Furthermore, we explore the effectiveness of enhancing students through knowledge injection training (RQ2) to reduce the frequency of teacher involvements.}


\section{\textsc{HaloCheck}}
A reliable model can respond to a given prompt in different phrasing while retaining the main information consistent with each other. We design \textsc{HaloCheck} to compute a  score $\mu$ for a set of sample responses $\mathbb{A} = \{A_1,..,A_n\}$, where $n$ is the number of sampled responses for a given prompt. In our experiments we set $n=5$ (chosen arbitrarily \footnote{High $n$ values can be time consuming.}).  The score $\mu$ indicates the level of consistency among the responses, where $min(\mu) \implies inconsistent(\mathbb{A})$ and $max(\mu) \implies consistent(\mathbb{A})$. 
Unlike selfcheckGPT \cite{manakul2023selfcheckGPT}, which calculates consistency between samples using sentence-level BERTScore or Question Answering (QA) agreement, our approach relies on sentence-level entailment between responses pairs, specifically utilizing \textsc{SummaC} \cite{laban-etal-2022-summac}, which was originally designed to measure the consistency between source text and generated summaries. In particular, we employ zero-shot sentence-level \textsc{SummaC} (\textsc{$SummaC_{zs}$}), utilizing a RoBERTa-based multilingual entailment model \cite{N18-1101}, to calculate the pairwise entailment between sentence pairs of two given LLM responses. The resulting score falls within the range of $[-1,1]$, where a score range between $[-1,0)$ indicates contradiction between the two texts ($-1$ indicates a total contradiction between samples), and a score range of $(0,1]$ suggests consistency ($1$ refers to perfect alignment). Furthermore, the original incentive of selfcheckGPT primarily aimed to assess the factuality of the generated response by gauging the level of sentence-level support based on other samples. In contrast, our approach places equal emphasis on all generated samples during evaluation, ensuring that the information provided remains consistent across all responses.

The rationale for adopting entailment is that responses might appear similar but still have contradictory information, potentially misleading selfcheckGPT-BERTScore and selfcheckGPT-ngram to inaccurately judge them as consistent. Additionally, for responses to questions, especially short ones, the use of n-gram models for probability estimation might be ineffective. 
Moreover, our approach removes the need for a question generation module like in selfcheckGPT-QA,  making our method more time efficient in estimating hallucination severity. We also conducted a quantitative and qualitative analysis to selfcheckGPT-NLI (which also utilized entailment methods  and wasn't included in the initial paper). We illustrate that  \textsc{HaloCheck} is more time efficient and more sensitive in detecting subtle contradictions within sampled responses \footnote{ Appendix \ref{sec:halocheck_examples} contains \textsc{HaloCheck} examples overcoming selfcheckGPT limitations. Appendix \ref{app:efficency} includes the evaluation time per 100 examples, highlighting algorithm efficiency.}.  

Pairwise entailment between responses can be represented as a bipartite graph such that its adjacency matrix $M$ can be expressed as a square matrix:

  $$ M = \begin{bmatrix}
    0 &  E^{T} \\
    E & 0 \\
  \end{bmatrix} $$


In the given equation, $U(M) = L(M)^{T}$, where $U$ and $L$ represent the upper and lower triangular matrices, respectively. This symmetry holds because $\textsc{SummaC}(A_i, A_j) = \textsc{SummaC}(A_j, A_i)$, where $i$ and $j$ represent the rows and columns in matrix $M$, respectively.
Thus, $E$ and $E^{T}$ in the matrix $M$ represent the \textsc{SummaC} scores computed between all combination pairs of the generated responses $A_1$ to $A_n$. Consequently, we compute the consistency score $\mu = mean(E)$. By building upon \textsc{$SummaC_{zs}$}, we conclude that the consistency score $\mu$ adheres to the range of $[-1,1]$, with a score of $-1$ indicating total contradiction between responses and a value of $1$ signifying complete alignment or agreement ($0$ is neutral).


\section{Knowledge injection}
\label{sec:k_i}
Scaling parameters of large pretrained language models can enhance their capacity to memorize world knowledge \cite{roberts-etal-2020-much}. Since we rely on LLMs with fewer parameters, we explore techniques to improve the model's ability to retain knowledge for domain-specific downstream tasks.

\subsection{Representing knowledge}
Our objective is to inject factual knowledge into LLMs without relying on scaling up instruction tuning datasets during supervised fine-tuning (SFT). Prior work by \citet{santos2022knowledge, moiseev2022skill} explored knowledge injection through prompt tuning of T5 \cite{raffel2020exploring} using entity triplets extracted from WikiData \cite{vrandevcic2014wikidata}. In a similar vein, we explore two types of knowledge representations: \textit{entity triplets} and \textit{entity summaries}. \textit{In this work, we narrow our focus to the NBA domain to simplify knowledge injection training and evaluation. Our intention is to expand our approach to encompass multiple domains in future research.}

\paragraph{Entity summaries}
We utilize the NBA entity list extracted from the NBA Wikipedia page \footnote{\url{https://en.wikipedia.org/wiki/National_Basketball_Association}} using the Wikipedia API \footnote{\url{https://pypi.org/project/Wikipedia-API/}}, which yields 1121 unique entities. For each entity in the list, we extract a one-page and two-page summary that serves as a  source  of trustworthy factual information. 

\paragraph{Entity triplets}
Each entity is treated as a subject entity from which we derive the corresponding object entity and the relationship between them using the WikiData API \footnote{\url{https://pypi.org/project/Wikidata/}}. For instance, for the entity \texttt{"Nikola Jokic"}, an example of an entity triplet would be \texttt{"Nikola Jokic, drafted by, Denver Nuggets"}. This approach unearths deeper knowledge beyond summaries. 
Unlike previous work \cite{sun2021ernie}, we preserved the triplet format instead of converting them into natural language. Remarkably, this decision didn't compromise the naturalness of the generated responses.

\textit{The final knowledge injection set is composed of $54K$ training samples.}

\subsection{Knowledge injection training}
\label{subsec:ki_train}
\citet{santos2022knowledge, moiseev2022skill} proposed employing a masked language model objective \cite{kenton2019bert} to predict the object entity based on the subject entity and the relation token. However, as our model is a decoder-only architecture, we maintain the same causal language model objective utilized in BLOOM pretraining and SFT \footnote{
We used a diverse in-house curated SFT dataset.}. To adapt our injection task to fit the finetuning objective, we introduced a special token, denoted as \texttt{"TRUE\_FACT:"} \footnote{We include an example of KI training in Appendix \ref{app:train_example}}, and prepended it to the knowledge text. During inference, this token can be leveraged to guide the model towards aligning with the factual knowledge witnessed during injection training \footnote{Training details included in Appendix \ref{app:train_inf}}.  We experiment with knowledge injection in two settings; $(1)$ \textit{\textbf{Intermediate tuning (interm. KI):}} in this setting, we first perform finetuning exclusively on the knowledge text. Afterwards, we further finetune the model on the SFT data. $(2)$ \textit{\textbf{Combined tuning (combine KI):}} where we combine knowledge injection text along with SFT data and jointly finetune the model. By conducting experiments with both settings, we aim to obtain a more comprehensive evaluation of the effectiveness of both techniques. Additionally, we intend to assess whether knowledge is retained or forgotten during the intermediate finetuning stage.

\section{Improving weaker LLM's consistency with a stronger teacher}
\label{sec:teacher_student}


We explore how stronger teachers, specifically GPT-4, can guide weaker student models (BLOOM7B) in reducing hallucinations during inference. We employed BLOOM7B  before and after knowledge injection (section \ref{sec:k_i})
to assess whether both techniques can complement each other. While augmented retrieval methods are a valid alternative, we focus on LLM teachers due to their ability to offer personalized answers and engage in multi-round conversations, which can be beneficial if hallucination persists after the initial round (although we only investigate single-round answers in this work, leaving multiple rounds for potential future research).

\subsection{Prompting teacher to provide the answer}

To assess if providing the student with the answer can reduce hallucinations, we assume the student consistently relies on the teacher's help, which allows us to estimate the upper bound of the consistent teacher's impact on improving the student's responses. We prompt GPT-4 accordingly, assigning it a meaningful role as an NBA expert and instructing it to answer questions accurately and faithfully. To provide detailed answers to the student model, we augment the prompt with automatic chain of thought (auto-CoT) \cite{zhang2022automatic}, prompting GPT-4 to break down its response to detailed steps. The prompt is structured as follows:

\begin{itemize}
    \item Role: \textit{"""You are an NBA expert. You will be given a question and your job is to answer accurately and faithfully."""}
    \item Question: \textit{"""question: \{\}"""}
    \item auto-CoT: \textit{"""answer step by step:"""}
\end{itemize}

The generated detailed answer is prepended to the question prior to prompting the student model.  

\subsection{Evaluate, then ask the teacher}

We utilize \textsc{HaloCheck} to trigger teacher answers only when inconsistencies are detected in the student's responses, as consistently relying on the teacher is impractical. Additionally, we investigate whether employing student models after knowledge injection can reduce costly teacher calls. Our goal is to find ways to optimize the teacher's guidance for the student's improvement while minimizing its usage.

 We conduct experiments with different \textsc{HaloCheck} thresholds within the consistency range of $0 < \text{threshold} < 1$, specifically using values $\{0, 0.2, 0.4, 0.6\}$ \footnote{
The chosen threshold ranges are designed to explore a variety of hallucination severities.}. The teacher is triggered if \textsc{HaloCheck} < threshold (indicating hallucinated response). We report the reduction in teacher calls as well as the improvement in \textsc{HaloCheck} scores for various hallucination thresholds.
\section{Question answering (QA) based evaluation}



Narrowing our scope to the NBA domain, we chose not to use existing factuality benchmarks like TruthfulQA \cite{lin2022truthfulqa} and Fact-Checking benchmarks \cite{thorne2018fever}, which cover broader domains and require reasoning abilities. Instead, we curated a domain-specific question-answering dataset focused on factual knowledge. This dataset forms the foundation for evaluating both \textsc{HaloCheck} and our hallucination reduction strategies.

\subsection{Developing a domain-specific Question Answering (QA) Dataset}
\label{subsec:qa-nba}

First, we utilize LLMs, specifically GPT-4 \cite{bubeck2023sparks}, to generate questions based on NBA-related entities and relevant background knowledge. This automated step helps generate a substantial pool of questions. Second, we manually review and filter the generated questions, retaining the most meaningful and relevant ones to ensure the dataset's quality for proper evaluation of our methods. 
\paragraph{Generating entity-based questions}

To generate relevant questions, we follow a systematic process. Initially, we randomly select 100 entities from the NBA entity list (as outlined in Section \ref{sec:k_i}). For each entity, we retrieve detailed information in the form of one-page summaries and entity triplets. Using GPT-4, we create questions based on this knowledge. Additionally, we utilize auto-CoT prompting to aid the model in breaking down the question generation process into manageable steps, enhancing the model's capability to produce well-structured and contextually appropriate questions. Examples of the generated questions are available in Appendix \ref{app:nba_qa_examples}. Our GPT-4 prompt consists of three components \footnote{Details about the parameters used are provided in Appendix \ref{app:qa_hyperparam}.}.
\begin{itemize}
    \item Role: \textit{"""You are an expert in NBA information, you will get an entity related to the NBA and related ground truth information in the format of either a  summary or a list of triplet relations. Your task is to generate a question related to truthful information and a short truthful answer. Keep your questions and answers simple."""}
    \item Entity and ground truth: \textit{"""entity: \{\} ground truth : \{\}"""}
    \item auto-CoT: \textit{"""Generate step by step:"""}
\end{itemize}

\paragraph{Filtering generated questions}
The initial generation process resulted in approximately 250 questions. To ensure the quality of the question list, we apply two filtering criteria: (1) Questions must be relevant to the NBA entities. (2) Questions can't serve as follow-up questions to previous ones, as BLOOM is a single-turn LLM and does not engage in multi-turn conversations. By applying these filtering conditions, we create a refined set composed of a $151$ question list that aligns with the requirements of the domain and BLOOM's single-turn context.

\subsection{Annotating QA generated answers}
\label{sec:annotation}

To evaluate the reliability of the \textsc{HaLoCheck} framework in estimating hallucination severity, we annotated a set of responses comprising 500 samples from five diverse BLOOM7B settings. Each setting consisted of 100 question responses. Specifically, we utilized answers from the following BLOOM7B settings: (1) \textit{BLOOM7B + SFT (baseline model)}, (2) \textit{BLOOM7B + SFT + teacher answers (GPT+CoT)}, (3) \textit{BLOOM7B + intermediate KI (KI includes both summary and entity triplets)}, (4) \textit{BLOOM7B + combined KI (KI includes both summary and entity triplets)}, and (5) \textit{BLOOM7B + combined KI (KI includes both summary and entity triplets) + teacher answers (GPT+CoT)}.

We conduct two types of annotation on the generated answer sets. Firstly, we assess the level of agreement between the samples in each answer set, which comprises 5 samples. This evaluation is termed \textit{\textbf{consistency}} and focuses on how closely the sampled answers align with one another. Secondly, we examine the \textit{\textbf{factuality}} of the sampled answers by comparing them to the ground truth. This allows us to evaluate the model's ability to consistently generate factual answers. We distinguish between \textit{consistency} and \textit{factuality} as the model can either provide consistent but inaccurate answers or give different accurate answers to the same question.

\paragraph{Consistency annotation} 
We split the task equally between two in-house annotators who rated each answer set on a 5-point Likert scale. The scale's values are as follows: $1:$ Indicates that none of the 5 samples agree with each other, implying a lack of consensus. $2:$ Two of the 5 samples agree with each other, showcasing some level of similarity. $3:$ Three of the 5 samples agree with each other, indicating moderate agreement. $4:$ Four of the 5 samples agree with each other, demonstrating a higher degree of consensus. $5:$ All 5 samples agree with each other, implying a strong alignment of information. 
We instruct annotators to assess agreement based on the conveyance of the same information rather than relying solely on identical wording.

\paragraph{Factuality annotation}
Following the guidelines introduced in \cite{manakul2023selfcheckGPT}, for each set of answers, we evaluate each sample  using a 3-point Likert scale: $1:$ a response that is unfactual. $2:$ a partially factual response, containing both factual and non-factual information. $3:$ a response that is factual and aligns with reality. To compute the final score for each answer set, we take the mean over the factuality of the $5$ responses generated by the LLM. Similar to the consistency annotation process, we involve two in-house annotators in this evaluation. Since the LLM's answers may not always be entirely factual, we provide the annotators with truthful background information (the Wikipedia page of the entities mentioned in the question) to cross-check the generated responses against it. This helps ensure accurate and reliable factuality assessments for each response.


\section{Results and Discussion}
\subsection{Correlation with Human Annotations}
\label{subsec:corr_human}
Table \ref{tab:corr_coeff} presents the correlation analysis, showcasing that \textsc{HaloCheck} exhibits higher correlations with human annotation scores for both consistency and factuality metrics compared to all selfcheckGPT approaches, including the recently introduced entailment-based ones. Notably, entailment-based methods perform better than those relying on BERTScore or QA methods, aligning with our hypothesis that entailment-based approaches can more effectively evaluate the severity of hallucinations.
To maintain consistency with \textsc{HaloCheck}, we report $1-$selfcheckGPT scores, as selfcheckGPT scores range from $[0,1]$ , where $0$ indicates perfect consistency.  The correlation with factuality scores is consistently lower than consistency scores across all metrics, possibly due to BlackBox metric limitations in capturing cases where the model consistently generates inaccurate responses. 
\begin{table}[]
\small
\begin{tabular}
{p{0.166\textwidth}|p{0.05\textwidth}p{0.05\textwidth}|p{0.05\textwidth}p{0.05\textwidth}}
\hline
\multirow{2}{*}{Metric}  & \multicolumn{2}{c}{Consistency} & \multicolumn{2}{c}{Factuality}  \\ \cline{2-5} 

                 & $\rho$  & $\tau$   & $\rho$   & $\tau$ \\ \hline
1-selfcheckGPT-BS    & 0.52                                                                                  &  0.43                                                                                    & 0.38                                                                                & 0.30                                                                                    \\ 
1-selfcheckGPT-QA    & 0.57                                                                                  & 0.45                                                                                     & 0.48                                                                                & 0.37   
\\
1-selfcheckGPT-NLI    & 0.73                                                                                 & 0.58                                                                                    & 0.61                                                                               &  0.47 \\
\hline

\textbf\textsc{HaloCheck} & \textbf{0.82}                                                                         & \textbf{0.68}                                                                            & \textbf{0.67}                                                                        & \textbf{0.53}                                                                           \\ \hline
\end{tabular}

\caption{Pearson ($\rho$) and Kendall tau ($\tau$) correlation coefficients with annotated Consistency and Factuality. \textbf{Bolded} indicates the most correlated. All scores are statistically significant with $p<0.05$. \label{tab:corr_coeff} }

\end{table}

\subsection{Evaluating knowledge injection strategies}

We present the average values of \textsc{HaloCheck} scores and selfcheckGPT variants across the entire QA dataset. We compare BLOOM7B with knowledge injection tuning to BLOOM7B finetuned on SFT data only and GPT-4. 

\begin{table}[ht]
\small
\begin{tabular}{p{0.15\textwidth}|p{0.11\textwidth}|p{0.031\textwidth}|p{0.031\textwidth}|p{0.031\textwidth}}
\hline
\multirow{2}{*}{Model}                           & \multirow{2}{*}{\textsc{HaloCheck}$\uparrow$}& \multicolumn{3}{c}{\centering selfcheckGPT$\downarrow$}  \\ \cline{3-5}
& & BS & QA & NLI\\ \hline
GPT-4                           &    0.72       &           0.14      &  0.26 &  0.17             \\ \hline
BLOOM7B+SFT                     & -0.38     & 0.58           & 0.52         &  0.87\\ 
+ interm. KI (summary)          & -0.22*     &          0.55*       & 0.47*       & 0.80*    \\ 
+ interm. KI (summary+triplets) & \textbf{-0.11*}     &       \textbf{0.44*}          &    0.41* & 0.69*            \\ 
+ combine KI (summary)          & -0.18*     & 0.48*          & 0.43*       &  0.73*   \\ 
+ combine KI (summary+triplets) &  -0.12*         & 0.46*                &          \textbf{0.38*}    & \textbf{0.67*}   \\ \hline
\end{tabular}

\caption{Average hallucination scores for QA answers. $\uparrow$ and $\downarrow$ indicate higher and lower is better respectively. \textbf{Bolded} numbers indicate the best performance compared to the BLOOM7B+SFT baseline. * indicates statistical significance with paired t-test compared to the BLOOM7B+SFT baseline. "interm." refers to intermediate tuning. \label{tab:k_i_res}}
\end{table}

Table \ref{tab:k_i_res} shows that all knowledge injection training methods significantly outperform BLOOM7B finetuned on SFT data alone across all evaluation metrics \footnote{Examples of knowledge injection effect is shown in Appendix \ref{app:k_i_effect}.}. Combining both summaries and triplets achieved the best results, confirming our hypothesis that triplets can uncover deeper relations and knowledge compared to using summaries alone. Additionally, it is worth noting that the performance of intermediate knowledge injection training aligns with that of combined knowledge injection training, suggesting that the model retains the injected knowledge during intermediate tuning. Despite the improvements achieved through knowledge injection, the model's performance remains inconsistent on average. This suggests that knowledge injection alone may not be sufficient to completely solve the hallucination problem, even when injecting domain-specific knowledge. 

\subsection{Improving BLOOM answers with teacher}

\begin{adjustwidth}{-\columnsep}{-\columnsep} 
\begin{figure*}[ht]
\small
  \centering
  \includegraphics[width=1.0\textwidth]{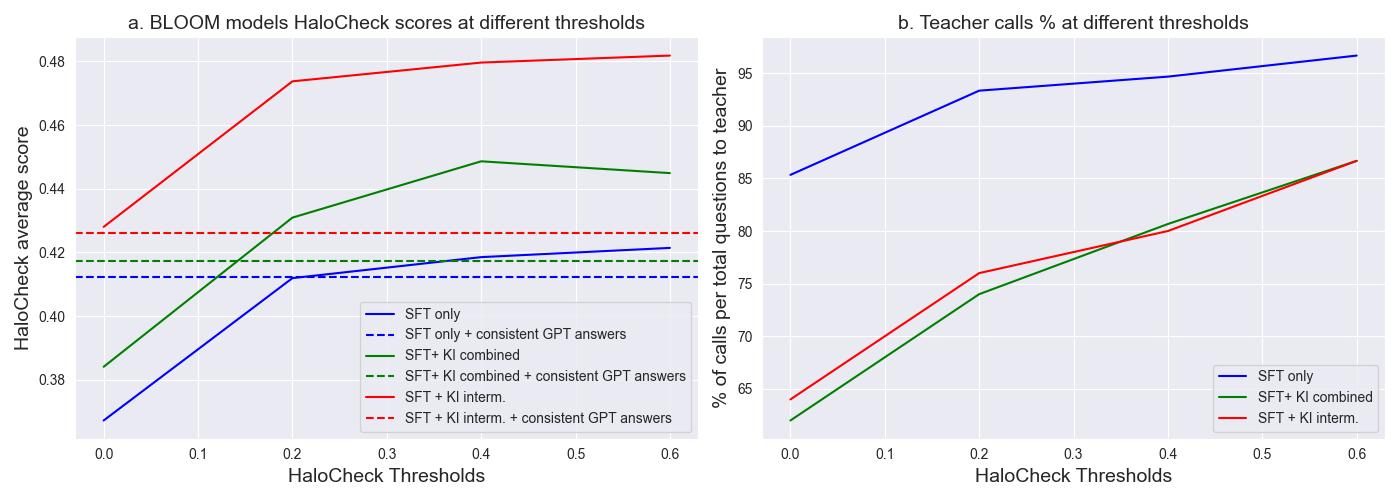} 
 \caption{\label{fig:teacher_student_analysis} (a) Illustrates the enhancement in \textsc{HaloCheck} based on varying the threshold across different levels of inconsistency. The $--$ lines indicate calling the teacher consistently for every question. (b) Shows the percentage of calls made to the teacher model in relation to the total number of questions, at different threshold levels.}
\end{figure*}
\end{adjustwidth}
Figure \ref{fig:teacher_student_analysis} a. illustrates the effectiveness of incorporating teacher-generated answers into student models. Integrating teacher answers consistently yields comparable \textsc{HaloCheck} scores of $0.42$ and $0.43$ for intermediate and combined fine-tuning scenarios (indicated by green and red dashed lines), respectively, compared to a baseline of $0.41$ (blue dashed line) for SFT-only model. The impact of knowledge injection becomes clearer at different call thresholds. Models benefiting from knowledge injection (SFT+KI interm. and SFT+KI combined) consistently outperform SFT-only models across hallucination thresholds. Intriguingly, restricting teacher answers to instances with higher hallucination severity further enhances model consistency, surpassing models that constantly rely on integrating teacher answers. These findings underscore the synergistic effect of the teacher-student approach and knowledge injection, ultimately improving overall model consistency. It's important to note that the degree of improvement diminishes beyond the $0.2$ hallucination threshold, suggesting a reduced benefit for higher-scoring answers through teacher-generated responses.


Figure \ref{fig:teacher_student_analysis}b. shows that knowledge injection effectively reduces the need for teacher interventions compared to the model without it. However, even with knowledge injection, $\sim 65\%$ of questions still require teacher assistance at the $0$ threshold. This continued reliance on the teacher highlights the challenge posed by the base model's inconsistency. This observation emphasizes the importance of exploring more robust models in future research to address this reliance and minimize teacher involvement.

\paragraph{Impact of teacher answers} Teacher answers can positively impact the model's consistency as illustrated in figure \ref{fig:teacher_student_analysis}. However, aligning with the findings in the reasoning task \cite{saha2023can}, we identified that inaccurate teacher answers can still mislead weaker student models, even after knowledge injection. Table \ref{tab:teacher_influence} demonstrates that when the teacher provides an incorrect answer, the student tends to follow suit with an erroneous response. Additionally, the table reveals that the use of auto-CoT can assist in mitigating these errors by enabling the teacher to offer more detailed information \footnote{Appendix \ref{app:gpt_effect} shows that auto-CoT effect isn't reflected quantitatively.}. 

\begin{table}[ht]
\small
\begin{tabular}{p{0.2\textwidth}|p{0.22\textwidth}}
\hline
\multicolumn{2}{c}{\textbf{Question:} When was LNB Pro A founded?}                                                                                                                                                                                                                                                                                                                                                                                                                                                                                                                                           \\ \hline
\textbf{GPT-4 (no auto-CoT):} & \textbf{GPT-4 (auto-CoT):} \\ \hline  "LNB Pro A is the top-tier men's professional basketball league in France. It was founded in 1987, and currently consists of 18 teams." &  1. LNB Pro A is the top-tier men's professional basketball league in France.

2. The league was founded in 1921 under the name "Championnat de France de Basket-Ball".

3. It was later renamed to "Ligue Nationale de Basket" (LNB) in 1987. 

4. The current name "LNB Pro A" was adopted in 2002. \\ \hline
BLOOM: \textit{1987}    \xmark                                                                                                                                                                 & BLOOM: \textit{1921}   \cmark                                                                                                                                                                                                                                                                                                                                     \\ \hline
\end{tabular}
\caption{Example of the effect of teacher's answers on student (BLOOM7B+ interm. KI) answers. \label{tab:teacher_influence}}
\end{table}


\section{Conclusion and future work}
This paper delved into the issue of quantifying and mitigating hallucinations in open-source weaker language models, using BLOOM7B as an example. To address the measurement of hallucinations \textbf{(RQ1)}, we introduced a novel BlackBox knowledge-free framework called \textsc{HaloCheck}. This framework not only outperformed similar methods in gauging the severity of hallucinations in language model outputs but also demonstrated notable efficiency in providing a fast overall performance estimate of hallucinations. To mitigate hallucinations, we explored knowledge injection into LLMs \textbf{(RQ2)}. Our findings indicate that this approach provides a straightforward means to enhance the models' domain-specific knowledge. Additionally, we examined the efficacy of leveraging a stronger teacher model \textbf{(RQ3)} to guide the weaker LLM. The results demonstrated the complementarity of this technique with knowledge injection, leading to improved performance of the less robust LLM. In the future, we intend to incorporate more robust student models while expanding our benchmark to encompass a broader range of tasks beyond domain-specific question answering.


\section*{Limitations}
In this work, we incorporated just one example of a weak open-source LLM (BLOOM7B). Furthermore, we intentionally focused exclusively on the NBA domain to facilitate our analysis. Additionally, depending on automatically generated questions has it's limitations, as these questions may not always align with human queries or may lack clarity in certain instances. Although our techniques can potentially extend to similar models and cover more domains, it's crucial to acknowledge that outcomes may vary. We encourage  researchers to expand upon our work to yield more comprehensive insights.

\section*{Ethics Statement}
Our primary focus is on addressing the measurement and reduction of hallucinations in LLMs. However, our experiments highlight that relying on these models, particularly weaker ones (as BLOOM7B), for obtaining critical and reliable information remains an ongoing challenge.

\bibliography{anthology,custom}
\bibliographystyle{acl_natbib}

\appendix

\section{Analyzing \textsc{HaloCheck} capabilities}
\subsection{Examples of \textsc{HaloCheck} capabilities in overcoming selfcheckGPT shortcomings}
\label{sec:halocheck_examples}

Table \ref{tab:halo_ex} provides three examples demonstrating \textsc{HaloCheck}'s superior ability in identifying the consistency of samples. In Example 1, where the sampled answers exhibit inconsistency, both \textsc{HaloCheck}, selfcheckGPT-QA, and selfcheck-NLI recognize this disparity. However, due to the high semantic similarity among the samples, selfcheckGPT-BS fails to align with other metrics.

In Example 2, the varying phrasing of sampled answers, coupled with differences in length, perplexes all selfcheckGPT metrics in capturing consistency. The top-ranked sampled answer offers more detail, unsupported by other samples, thus misleading selfcheckGPT metrics. Conversely, \textsc{HaloCheck} computes a consistent score by considering pairwise sentence similarity across all samples, aligning better with the correctness of samples.

Our metric proves more effective at detecting subtle contradictions among samples compared to selfcheckGPT-NLI, as evident in Example 3. While selfcheckGPT-NLI assigns a near-perfect score close to 1 (indicating strong contradiction), \textsc{HaloCheck} assigns a moderate agreement score greater than 0 (indicating agreement, though not perfect). This discrepancy in scoring corresponds with the samples' consensus on the Los Angeles Lakers facing the New York Knicks in the final game, while differing on details such as the number of games played and final scores.
\begin{table*}[]
\small
\centering
\begin{tabular}{p{0.5\textwidth}|p{0.11\textwidth}|p{0.07\textwidth}|p{0.07\textwidth}|p{0.07\textwidth}}
\hline
\multirow{2}{*}{ Examples}                           & \multirow{2}{*}{\textsc{HaloCheck}$\uparrow$}& \multicolumn{3}{c}{\centering selfCheckGPT$\downarrow$}  \\ \cline{3-5}
& & BS & QA & NLI \\ \hline
\begin{center}
    \textbf{Question:} When was LNB Pro A founded? \\

    sample1: LNB Pro A was founded in \textbf{1979.}\\
    sample2: LNB Pro A was founded in \textbf{1988.}\\
    sample3: LNB Pro A was founded in \textbf{1989.}\\
    sample4: LNB Pro A was founded in \textbf{1998.} \\
  sample5: LNB Pro A was founded in \textbf{1983.}
\end{center}

& \multirow{6}{*}{-0.91}   & \multirow{6}{*}{0.0004}          & \multirow{6}{*}{0.85}   & \multirow{6}{*}{0.99}           \\ \hline

\begin{center}
    \textbf{Question:} When was the 1958 NBA Finals played? \\

   Sample1: The 1958 NBA Finals was played in April. The St. Louis Hawks won the 1958 NBA Finals\\ 
   Sample2: The 1958 NBA Finals was played from April 1-13, 1958.\\
   Sample3:  April 1-13, 1958.\\
   Sample4:  April 1-13, 1958.\\
   Sample5:  April 1-13, 1958.\\
\end{center}

& \multirow{8}{*}{0.73}   & \multirow{8}{*}{1.06}          & \multirow{8}{*}{0.62}        & \multirow{8}{*}{0.87}       \\ \hline

\begin{center}
     \textbf{Question:}  Which teams played in the 1972 NBA Finals? \\

    $\bullet$ \textbf{sample1:} The 1972 NBA Finals was played between the Los Angeles Lakers and New York Knicks. The Lakers won the series 4 games to 1.\\
    $\bullet$ \textbf{sample2:} The 1972 NBA Finals was played between the Los Angeles Lakers and New York Knicks. The Lakers won the series 4 games to 2 for their second championship.\\
    $\bullet$ \textbf{sample3:} The 1972 NBA Finals was played between the Los Angeles Lakers and New York Knicks. The Lakers won the series 4 games to 2 for their second championship.\\
    $\bullet$ \textbf{sample4:} The 1972 NBA Finals was played between the Los Angeles Lakers and New York Knicks. The Lakers won the series 4 games to 2 for their second consecutive NBA title. \\
    $\bullet$ \textbf{sample5:} The 1972 NBA Finals was played on June 17, 1971. The Western Conference champion Los Angeles Lakers defeated the Eastern Conference champion New York Knicks in five games to win their sixth title.: Knicks 50–LAN 37, Knicks 41–LAN 48, Knicks 40–LAN 57, Knicks 52–LAN 80, Knicks 86–LAN 105. 
         
     \end{center}

& \multirow{20}{*}{0.11} & \multirow{20}{*}{0.34} & \multirow{20}{*}{0.35}  & \multirow{20}{*}{0.99}              \\ \hline

\end{tabular}

\caption{\textsc{HaloCheck} examples in overcoming selfCheckGPT \label{tab:halo_ex}}
\end{table*}

\begin{table}[ht]
\small
\begin{tabular}{p{0.1\textwidth}|p{0.09\textwidth}|p{0.09\textwidth}|p{0.09\textwidth}}
\hline
\multicolumn{4}{c}{ \centering Average time taken in seconds/100 examples}          \\ \hline
\multirow{2}{*}{\textsc{HaloCheck}} & \multicolumn{3}{c}{ selfcheckGPT} \\ \cline{2-4} 
   & QA & BS   & NLI    \\ \hline
\textbf{54.22}                       &  956.54  & 507.57 &  135.00 \\ \hline
\end{tabular}
\caption{Average time in seconds per 100 examples taken to compute each metric. \textbf{Bold} indicates the lowest time.   \label{tab:avg_time}}
\end{table}

\subsection{\textsc{HaloCheck} time efficiency}
\label{app:efficency}
Table \ref{tab:avg_time} illustrates that \textsc{HaloCheck} consumes considerably less time compared to selfCheckGPT methods. This efficiency not only makes \textsc{HaloCheck} superior in hallucination estimation, but also provides a fast and swift performance estimate. To compute the average time taken, we randomly sample $100$ question responses $5$ times and compute the average. Time estimates was done on an \textit{V-100 Tesla GPU}.

\section{NBA Question Answering}
\label{app:nba_qa}

\begin{table}[]
\small
\begin{tabular}{p{0.11\textwidth}|p{0.3\textwidth}}
\hline
Entities          & Questions                                   \\ \hline
Adam Silver       & $1.$ Who is the current commissioner of the NBA? \\ \hline
The NBA MVP Award & $2.$ How is the MVP award winner determined?     \\ \hline
Maurice Podoloff  & $3.$ Who is Maurice Podoloff?                    \\ \hline
\end{tabular}
\caption{Examples of selected filtered questions generated from GPT-4 \label{tab:ques_gen}}
\end{table}

\subsection{Examples of the generated questions}
\label{app:nba_qa_examples}

Table \ref{tab:ques_gen} displays examples of the filtered generated questions by GPT-4. The questions vary in complexity, ranging from simple retrieval-type questions like $1.$, where the answer is a straightforward entity, to more descriptive questions like $3.$, which typically require answers composed of multiple sentences. Additionally, some questions, like $2.$, demand step-by-step explanations regarding specific entities. This diverse set of questions enables us to more effectively assess the model's hallucinations across different scenarios.

\subsection{Prompting details and hyperparameters for GPT-4}
\label{app:qa_hyperparam}

We utilized the OpenAI API \footnote{\url{https://openai.com/blog/openai-api}} to prompt GPT-4, with the \textit{temperature} set to $0.7$ and \textit{top\_p} (nucleus sampling \cite{holtzman2019curious}) to $0.3$ to encourage output diversity during question generation. From the generated responses, we extracted the top response containing questions and their corresponding answers based on the provided prompt and background knowledge for each entity. Since answers weren't always factual, we opted against using them for evaluation.

\section{Training details}

\subsection{Knowledge Injection training format}
\label{app:train_example}
Table \ref{tab:k_i_example_train} shows an example of the training samples used for knowledge injection training. In all knowledge injection instances, we replaced the instruction with "\texttt{TRUE\_FACT}" during instruction tuning.
 
\begin{table}[ht]
\small
\begin{tabular}{p{0.45\textwidth}}
\hline
Training examples \\ \hline
$1.$ \texttt{TRUE\_FACT:} Nikola Jokic drafted by Denver Nuggets   \\ \hline
$2.$ \texttt{
TRUE\_FACT:} Nikola Jokic (Serbian Cyrillic:  born February 19, 1995) is a Serbian professional basketball player who is a center for the Denver Nuggets of the National Basketball Association (NBA). Nicknamed \"the Joker\", and hailed as one of the biggest draft steals in NBA history, he is regarded as one of the greatest players and centers of all time. A five-time NBA All-Star, he has been named to the All-NBA Team on five occasions (including three first-team selections), and won the NBA Most Valuable Player Award for the 2020-21 and 2021-22 seasons. He represents the Serbian national team with which he won a silver medal at the 2016 Summer Olympics \\ \hline
             
\end{tabular}
\caption{Examples of knowledge injection training samples. Row $1.$ is knowledge injection with a triplet example and $2.$ is knowledge injection with summary example. \label{tab:k_i_example_train}}
\end{table}

\subsection{Training implementation}
\label{app:train_inf}
The BLOOM7B implementation used in our experiments is from the \textit{HuggingFace} library \cite{wolf2019huggingface}. Each sample is trained for $100$ epochs. In the case of \textit{intermediate knowledge injection} tuning, we perform further tuning on the best checkpoint that obtained the lowest loss.

\section{Knowledge injection effect}
\label{app:k_i_effect}

Table \ref{tab:k_i_answers} illustrates the impact of knowledge injection on BLOOM. Notably, in Example 1, the initial answers remained related to NBA teams, although not entirely unrelated. However, the model exhibited both inconsistency and inaccuracies in delivering factual responses. Following knowledge injection, we observe enhanced consistency. Remarkably, with combined knowledge injection, the model achieves flawless consistency and factuality, delivering correct answers in each sample. This underscores the potency of knowledge injection in elevating the model's performance.

In Example 2, the model maintained consistency both before and after knowledge injection. Nevertheless, an advantage of knowledge injection lies in updating outdated information. Before knowledge injection, the model provided the former name of the Golden State Warriors' home arena (The Oracle Arena). However, after knowledge injection, it correctly gives the updated arena name (The Chase Center). This showcases the value of knowledge injection in keeping the model's information up-to-date.

\begin{table*}[ht]
\small
\begin{tabular}{p{0.25\textwidth}|p{0.25\textwidth}|p{0.25\textwidth}|p{0.25\textwidth}}
\hline
Question                     & SFT only                                                                                                                                                                                                                                                                                                                                                & Intermediate  KI (summary + triplets)                                                                                                                                                                                                                                                                                                                                                                                                                                                                                                         & Combined KI (summary + triplet)                                                                                                                                                                                 \\ \hline
1. Who won the 1952 NBA Finals? & $\bullet$  sample1:  New York Knicks.

$\bullet$ sample2: The Boston Celtics won the 1952 NBA Finals against the New York Knicks, 4 games to 1. 

$\bullet$  sample3: New York Knicks.

$\bullet$  sample4: New York Knicks.

$\bullet$ sample5: The Boston Celtics defeated the New York Knicks 5 games to 3 to win their second championship in franchise history. & $\bullet$ sample1: The Minneapolis Lakers won the 1952 NBA Finals. 

$\bullet$ sample2: The Minneapolis Lakers won the 1952 NBA Finals.

$\bullet$ sample3: The Minneapolis Lakers faced the New York Knicks. The Minneapolis Lakers won the 1952 NBA Finals. 

$\bullet$ sample4: The Minneapolis Lakers faced the New York Knicks in a best-of-seven series with Minneapolis having home-court advantage.  

$\bullet$ sample5: The Minneapolis Lakers faced the New York Knicks in a best-of-seven series with Minneapolis having home-court advantage. & $\bullet$ sample1: The Minneapolis Lakers.

$\bullet$  sample2: The Minneapolis Lakers. 

$\bullet$ sample3: The Minneapolis Lakers.

$\bullet$ sample4: The Minneapolis Lakers.

$\bullet$ sample5: The Minneapolis Lakers\\ \hline

2. What is the Golden State Warriors' home arena? & $\bullet$ The Golden State Warriors' home arena is the Oracle Arena in Oakland, CA. 

$\bullet$ The Golden State Warriors' home arena is the Oracle Arena in Oakland, CA.

$\bullet$ The Golden State Warriors' home arena is the Oracle Arena in Oakland, CA.

$\bullet$ The Golden State Warriors' home arena is the Oracle Arena in Oakland, CA.

$\bullet$ The Golden State Warriors' home arena is the Oracle Arena in Oakland, California.

$\bullet$ The Golden State Warriors' home arena is the Oracle Arena in Oakland, California. & 

$\bullet$ The Golden State Warriors home arena is the Chase Center. 

$\bullet$ The Golden State Warriors home arena is the Chase Center.

$\bullet$ The Golden State Warriors home arena is the Chase Center.

$\bullet$ The Golden State Warriors home arena is the Chase Center.

$\bullet$ The Golden State Warriors home arena is the Chase Center. &

$\bullet$ The Golden State Warriors home arena is the Chase Center. 

$\bullet$ The Golden State Warriors home arena is the Chase Center.

$\bullet$ The Golden State Warriors home arena is the Chase Center.

$\bullet$ The Golden State Warriors home arena is the Chase Center.

$\bullet$ The Golden State Warriors home arena is the Chase Center.\\ \hline

\end{tabular}
\caption{Examples of answers improvement after knowledge injection. \label{tab:k_i_answers}}
\end{table*}

\section{GPT answers effect}
Table \ref{tab:auto-Cot-analysis} underscores the consistent advantages stemming from the inclusion of GPT-generated answers, both with and without auto-CoT. This practice yields improvements across all baselines. Moreover, the model's consistency can experience enhancements solely through the utilization of GPT-generated answers, surpassing the improvements achieved when answers are accompanied by auto-CoT. However, acknowledging the substantial influence teacher answers exert on student responses, as demonstrated in Table \ref{tab:teacher_influence}, we advocate for relying on GPT + auto-CoT as the teacher (as adopted in our teacher-student analysis). This strategy facilitates the organization of factual content within student responses, thereby enhancing the factual accuracy of the generated answers at the cost of a slight reduction in the consistency of the answers after integrating the teacher's detailed responses. Furthermore, we foresee the potential for auto-CoT to generalize and aid other hallucination-prone student models in tasks that demand more complex reasoning beyond our current knowledge-intensive question answering context.

We have also added an additional example to Table \ref{tab:teacher_inf_more_examples}, further reinforcing the observed trends in Table \ref{tab:teacher_influence}. As illustrated in the table, the inclusion of auto-CoT assists the model in breaking down the answers into more detailed components, thereby improving its ability to approach the correct generated response more effectively.

\label{app:gpt_effect}

\begin{table*}[ht]
\centering
\small
\centering
\begin{tabular}{p{0.45\textwidth}|p{0.11\textwidth}|p{0.031\textwidth}|p{0.031\textwidth}|p{0.031\textwidth}}
\hline
\multirow{2}{*}{Model}                           & \multirow{2}{*}{\textsc{HaloCheck}$\uparrow$}& \multicolumn{3}{c}{\centering selfcheckGPT$\downarrow$}  \\ \cline{3-5}
& & BS & QA & NLI\\ \hline

\textit{BLOOM7B+SFT}  & \textit{-0.38}     & \textit{0.58}           & \textit{0.52}         &  \textit{0.87}\\  \hline 

+ GPT                     &  0.53    &  0.24        &   0.19      &  0.27\\ 
+ GPT-CoT          & 0.41    &      0.37          &      0.25  & 0.37   \\ \hline
 \textit{BLOOM7B + SFT + interm. KI (summary+triplets)} & \textit{-0.11} & \textit{0.44} & \textit{0.41} & \textit{0.69} \\ \hline 

+ GPT & 0.51     &    0.24         & 0.18    &  0.29           \\ 
+  GPT-CoT & 0.43 & 0.35 & 0.29 & 0.41 \\ \hline 
 \textit{BLOOM7B + SFT+ combine KI (summary+triplets)} & \textit{-0.12} & \textit{0.46} & \textit{0.38} & \textit{0.67} \\ \hline

+ GPT &  0.50         &   0.24             &   0.19           &  0.28 \\ 

+ GPT-CoT &  0.42        &   0.35
&      0.27       &  0.41  \\ \hline
\end{tabular}

\caption{Average hallucination scores of using GPT answers with and without auto-CoT (CoT) consistently for the QA task. $\uparrow$ and $\downarrow$ indicate higher and lower is better respectively. \textit{Italicized} indicates no GPT answers employed.\label{tab:auto-Cot-analysis}}
\end{table*}

\begin{table*}[ht]
\centering
\small
\begin{tabular}{p{0.45\textwidth}|p{0.45\textwidth}}
\hline
\multicolumn{2}{c}{\textbf{Question:} Who won the 1958 NBA Finals?}                                                                                                                                                                                                                                                                                                                                                                                                                                                                                                                                           \\ \hline
\textbf{GPT-4 (no auto-CoT):} & \textbf{GPT-4 (auto-CoT):} \\ \hline  "The 1958 NBA Finals was played between the Boston Celtics and the St. Louis Hawks. The Celtics won the series 4-2 to claim their first NBA championship." &  The 1958 NBA Finals was won by the St. Louis Hawks. Here are some details about the series:

- The 1958 NBA Finals was a best-of-seven series between the St. Louis Hawks and the Boston Celtics.

- The series began on April 5, 1958 and ended on April 13, 1958.

- The Hawks won the series 4 games to 2.

- The series MVP was Bob Pettit of the St. Louis Hawks.

- This was the first and only NBA championship won by the St. Louis Hawks franchise.  \\ \hline

BLOOM: \textit{The Boston Celtics}    \xmark                                                                                                                                                                 & BLOOM: \textit{The St. Louis Hawks}   \cmark                                                                                                                                                                                                                                                                                                                                     \\ \hline
\end{tabular}

\caption{
    More examples of the effect of teacher's answers on student (BLOOM7B+ interm. KI) answers. All sampled answers were consistent and eliminated to save space. 
 \label{tab:teacher_inf_more_examples}}

\end{table*}
\end{document}